\documentclass[11pt]{article}

\usepackage[preprint]{acl}

\usepackage{times}
\usepackage{latexsym}

\usepackage[T1]{fontenc}

\usepackage[utf8]{inputenc}

\usepackage{microtype}

\usepackage{inconsolata}

\usepackage{graphicx}

\usepackage{amsmath}
\usepackage{amssymb}
\usepackage{booktabs}
\usepackage{multirow}
\usepackage{float}
\usepackage{subcaption}

\title{CEFR Level Assessment across Languages and Text Granularities}

\author{Eugene Matsumura \\
  RWTH Aachen University \\
  \texttt{eugene.matsumura@rwth-aachen.de} \\\And
  Yuki Arase \\
  Institute of Science Tokyo \\
  \texttt{arase@c.titech.ac.jp} \\}

\begin{document}
\maketitle
\begin{abstract}
Automatic assessment of textual CEFR levels is an important task in educational natural language processing. 
However, accurate CEFR level prediction remains challenging due to the naturally imbalanced distribution of CEFR levels and the ambiguous boundaries between adjacent levels. 
To address these challenges, we propose a unified training objective that combines focal loss with supervised contrastive learning. 
Focal loss mitigates class imbalance by emphasizing underrepresented and difficult examples, while supervised contrastive learning learns more discriminative representations by encouraging intra-class compactness and inter-class separation. 
We evaluate the proposed method on sentence- and document-level CEFR assessment in both monolingual and multilingual settings. 
Experimental results demonstrate that our method consistently outperforms strong baselines across all settings, establishing new state-of-the-art performance on multiple CEFR assessment benchmarks.
\end{abstract}

\section{Introduction}
Automatic text difficulty assessment has long been an active research topic in natural language processing, with applications in educational technologies such as automated essay scoring and text simplification \citep{li:2025}. 
The Common European Framework of Reference (CEFR)\footnote{\url{https://www.coe.int/en/web/common-european-framework-reference-languages}} is a widely adopted framework for describing language proficiency, defining six levels ranging from A1 (beginner) to C2 (proficient).

Early approaches to predicting CEFR levels of texts relied on manually engineered linguistic features \citep{vajjala-meurers:2012}, while more recent work \citep{imperial:2025} has substantially improved performance using pretrained language models such as BERT \cite{devlin:2019}. 
Despite these advances, two fundamental challenges remain.
First, the distribution of CEFR levels is inherently imbalanced, with A1 and C2 texts occurring much less frequently than texts at intermediate levels. Second, CEFR levels form an ordinal scale, and the boundaries between adjacent levels are often ambiguous, making neighboring levels difficult to distinguish consistently.
Although \citet{arase:2022} addressed class imbalance through metric learning, their approach does not explicitly account for the ambiguity of adjacent CEFR levels.

To address these challenges, we propose a CEFR level assessment method that jointly leverages focal loss \citep{lin:2017} and supervised contrastive learning \citep{gao:2021}. 
Focal loss mitigates the effect of class imbalance by emphasizing underrepresented and difficult examples, while supervised contrastive learning learns more discriminative representations by pulling together examples from the same CEFR level while increasing the separation between different CEFR levels in the embedding space. 
Together, these objectives improve CEFR level prediction under both class imbalance and ambiguous class boundaries.

We evaluate the proposed method on sentence- and document-level CEFR assessment in both monolingual (English) and multilingual settings. 
Experimental results demonstrate that it consistently outperforms strong state-of-the-art baselines across all evaluation settings.

The contributions of this paper are threefold: (i) we propose a unified training objective combining focal loss and supervised contrastive learning for CEFR level assessment, addressing both naturally imbalanced CEFR distributions and ambiguity between adjacent levels; (ii) we demonstrate its effectiveness on sentence- and document-level CEFR level assessment in both monolingual and multilingual settings; and (iii) we achieve state-of-the-art on multiple CEFR level assessment datasets.

\section{Proposed Method}
Following \citet{arase:2022}, we formulate CEFR level prediction as a classification task: predict a class label $y \in \{1, \dots, K\} (K=6)$ given an input text $X = (x_1, \dots, x_T)$. 
We employ a pretrained masked language model (PMLM) for encoding the input text. 
The hidden representation $\mathbf{z} \in \mathbb{R}^d$ is passed to a classification head, which is a single linear layer mapping $\mathbf{z}$ to $K$ logits, followed by a softmax that yields the class posterior.

We conduct two-stage training; as in \citet{khosla:2020}, preliminary experiments confirmed this separation performed better than joint training.
In the first stage, the PMLM parameters are fine-tuned with the supervised contrastive learning (Section~\ref{sec:contrastive_learning}). 
In the second stage, the first eight layers of the embedder are frozen, and the classification head is trained on top of it with the focal loss (Section~\ref{sec:focal_loss}). 

\subsection{Encoding}
First, the input text $X$ is encoded by $\operatorname{PMLM}$: $\mathbf{h}_1, \mathbf{h}_2,\ldots, \mathbf{h}_T = \operatorname{PMLM}(X)$,
where $\mathbf{h}_t \in \mathbb{R}^d$ is the final layer hidden representation. 
We then obtain a fixed-size sentence representation, $\mathbf{z}$, by mean pooling over the sequence dimension: $\mathbf{z} = \operatorname{MeanPool}(\mathbf{h}_1, \mathbf{h}_2, \ldots, \mathbf{h}_T).$

\subsection{Supervised Contrastive Learning}
\label{sec:contrastive_learning}
Supervised Contrastive Learning \cite{khosla:2020} regards samples sharing the same class label as positive pairs, while all remaining samples in the batch as negatives. 
Formally, given an $i$-th sample, the probability assigned to a positive pair $p$ is
\begin{align}
p_{\mathrm{con}}(i,p)
&=
\frac{\exp(\operatorname{sim}(\mathbf{z}_i,\mathbf{z}_p)/\tau)}
{\sum_{a\in A(i)}
\exp(\operatorname{sim}(\mathbf{z}_i,\mathbf{z}_a)/\tau)},\\
A(i)&=\{1, 2, \ldots, 2N \} \setminus \{ i\}
\end{align}
where $\operatorname{sim}(\cdot, \cdot)$ is cosine similarity measuring the closeness of two embeddings, $\tau$ is a temperature hyperparameter of the distribution, and $N$ is the number of training samples. 
We augment the samples by encoding each input text using the PMLM twice. 
The randomness of the dropout layers yields two slightly different embeddings of the same sentence, which serves as the augmented positive pair \cite{gao:2021}. 
Finally, the loss function is defined as:
\begin{align}
\mathcal{L}_{\text{SupCon}}
&=
\sum_{i=1}^{2N}
\frac{-1}{|P(i)|}
\sum_{p\in P(i)}
\log p_{\mathrm{con}}(i,p),\\
P(i) &= \{p \in A(i) \mid y_p = y_i \}.
\label{eq:supcon}
\end{align}

\subsection{Focal Loss}
\label{sec:focal_loss}
The focal loss \cite{lin:2017} is a generalization of the cross-entropy loss that introduces a modulating factor to adjust the contribution of each sample to the total loss.
For a training example with ground-truth class $y$, it is defined as
\begin{equation}
    \mathcal{L}_{\mathrm{focal}} = -(1 - p_y)^{\gamma} \log p_y ,
\end{equation}
where $p_y$ denotes the predicted probability of the ground-truth class and $\gamma \geq 0$ is the focusing parameter.
For $\gamma = 0$, the focal loss reduces to the standard cross-entropy loss and for $\gamma > 0$, it down-weights well-classified examples and shifts training towards hard ones.

Focal loss adaptively reweights individual training examples based on their prediction confidence. 
To additionally account for the global class distribution, we combine it with the class-level weighting scheme \cite{arase:2022}, which weights the loss for each CEFR level according to its frequency in the training data: $w_i = \frac{q_i^{\alpha}}{\sum_{j=1}^{K} q_j^{\alpha}}$, where $q_i$ is the frequency of level $i$ in the training set and $\alpha \in [0, 1]$ controls the weight strength.

\begin{table}[t]
\centering
\begin{tabular}{llr}
\toprule
Dataset & Split & Num. of samples \\
\midrule
CEFR-SP 
 & train & 13,990 \\
 & dev   & 1,843 \\
 & test  & 1,843 \\
\midrule
UCEFR-sent / -doc
 & train & 39,473 / 25,537 \\
 & dev   & 6,146 / 6,191 \\
 & test  & 1,400 / 1,241 \\
\bottomrule
\end{tabular}
\caption{Number of training, development, and test samples in CEFR-SP and UniversalCEFR.}
\label{tab:num_samples}
\end{table}


\begin{table*}[t]
\centering
\begin{tabular}{llccc}
\toprule
Dataset & CEFR level & Cross-Entropy & Prototype & SupCon + Focal \\
        &            &               & \citep{arase:2022} & (proposed) \\
\midrule
\multirow{7}{*}{CEFR-SP}
 & A1 & $0.319$ \tiny{$\pm0.257$} & $0.761$ \tiny{$\pm0.024$} & \underline{$0.779$} \tiny{$\pm0.018$} \\
 & A2 & $0.843$ \tiny{$\pm0.012$} & $0.822$ \tiny{$\pm0.021$} & \underline{$0.844$} \tiny{$\pm0.008$} \\
 & B1 & \underline{$0.876$} \tiny{$\pm0.013$} & $0.859$ \tiny{$\pm0.016$} & $0.871$ \tiny{$\pm0.010$} \\
 & B2 & $0.854$ \tiny{$\pm0.027$} & $0.856$ \tiny{$\pm0.029$} & \underline{$0.869$} \tiny{$\pm0.010$} \\
 & C1 & $0.827$ \tiny{$\pm0.033$} & \underline{$0.861$} \tiny{$\pm0.016$} & $0.856$ \tiny{$\pm0.014$} \\
 & C2 & $0.586$ \tiny{$\pm0.203$} & \underline{$0.884$} \tiny{$\pm0.038$} & $0.861$ \tiny{$\pm0.050$} \\
\cmidrule{2-5}
 & Macro-F1 & $0.718$ \tiny{$\pm0.024$} & $0.841$ \tiny{$\pm0.010$} & $\mathbf{0.847}$ \tiny{$\pm0.006$} \\
\midrule
\multirow{6}{*}{UCEFR-sent}
 & A1 & $0.783$ \tiny{$\pm0.023$} & $0.775$ \tiny{$\pm0.014$} & \underline{$0.833$} \tiny{$\pm0.009$} \\
 & A2 & $0.579$ \tiny{$\pm0.007$} & $0.555$ \tiny{$\pm0.006$} & \underline{$0.610$} \tiny{$\pm0.011$} \\
 & B1 & \underline{$0.584$} \tiny{$\pm0.025$} & $0.542$ \tiny{$\pm0.028$} & $0.548$ \tiny{$\pm0.003$} \\
 & B2 & \underline{$0.593$} \tiny{$\pm0.013$} & $0.561$ \tiny{$\pm0.025$} & $0.582$ \tiny{$\pm0.026$} \\
 & C  & $0.688$ \tiny{$\pm0.029$} & $0.696$ \tiny{$\pm0.003$} & \underline{$0.700$} \tiny{$\pm0.020$} \\
\cmidrule{2-5}
 & Macro-F1 & $0.646$ \tiny{$\pm0.010$} & $0.626$ \tiny{$\pm0.012$} & $\mathbf{0.655}$ \tiny{$\pm0.007$} \\
\midrule
\multirow{6}{*}{UCEFR-doc}
 & A1 & $0.794$ \tiny{$\pm0.032$} & $0.790$ \tiny{$\pm0.015$} & \underline{$0.808$} \tiny{$\pm0.005$} \\
 & A2 & $0.740$ \tiny{$\pm0.011$} & $0.720$ \tiny{$\pm0.030$} & \underline{$0.753$} \tiny{$\pm0.023$} \\
 & B1 & $0.706$ \tiny{$\pm0.015$} & $0.666$ \tiny{$\pm0.015$} & \underline{$0.709$} \tiny{$\pm0.011$} \\
 & B2 & $0.642$ \tiny{$\pm0.021$} & $0.643$ \tiny{$\pm0.030$} & \underline{$0.645$} \tiny{$\pm0.007$} \\
 & C  & $0.709$ \tiny{$\pm0.014$} & $0.692$ \tiny{$\pm0.032$} & \underline{$0.727$} \tiny{$\pm0.022$} \\
\cmidrule{2-5}
 & Macro-F1 & $0.718$ \tiny{$\pm0.006$} & $0.702$ \tiny{$\pm0.009$} & $\mathbf{0.728}$ \tiny{$\pm0.002$} \\
\bottomrule
\end{tabular}
\caption{Per-level F1 and macro-F1 on the CEFR-SP and UniversalCEFR (sentence- and document-level) test sets.
For CEFR-SP, we report mean~$\pm$~std over $12$ seeds, discarding the best- and worst-performing seed by macro-F1 (i.e., averaging over the remaining $10$); for UniversalCEFR, mean~$\pm$~std over $3$ seeds.
The \textbf{best Macro-F1 scores} are indicated by bold fonts, while the \underline{best per-level F1 scores} are indicated by underlines.
}
\label{tab:main-results}
\end{table*}

\section{Experiments}
We evaluate the performance of the proposed method on multi-granular and multilingual settings. 

\subsection{Datasets}
We use two datasets, CEFR-SP \cite{arase:2022}, an English sentence-level corpus, and UniversalCEFR \cite{imperial:2025}, a multilingual collection spanning multiple text granularities. 
Table~\ref{tab:num_samples} shows the number of samples in each dataset. 
For the label distributions of CEFR levels, please refer to Table~\ref{tab:dataset-distribution} in Appendix~\ref{app:dataset-stats}.

CEFR-SP provides labels assigned by two annotators. 
We expand our contrastive learning setting to regard both labels as correct CEFR levels (see Appendix~\ref{sec:dual_label} for more details).

UniversalCEFR covers $13$ languages; we pool all languages and conduct separate experiments at the sentence and document level (denoted as ``UCEFR-sent'' and ``UCEFR-doc'', respectively).
Because C2 samples are scarce on UniversalCEFR, we merge C1 and C2 into a single C class to discuss based on stable statistics.

\begin{figure*}[t!]
  \centering 
  \includegraphics[width=0.8\textwidth]{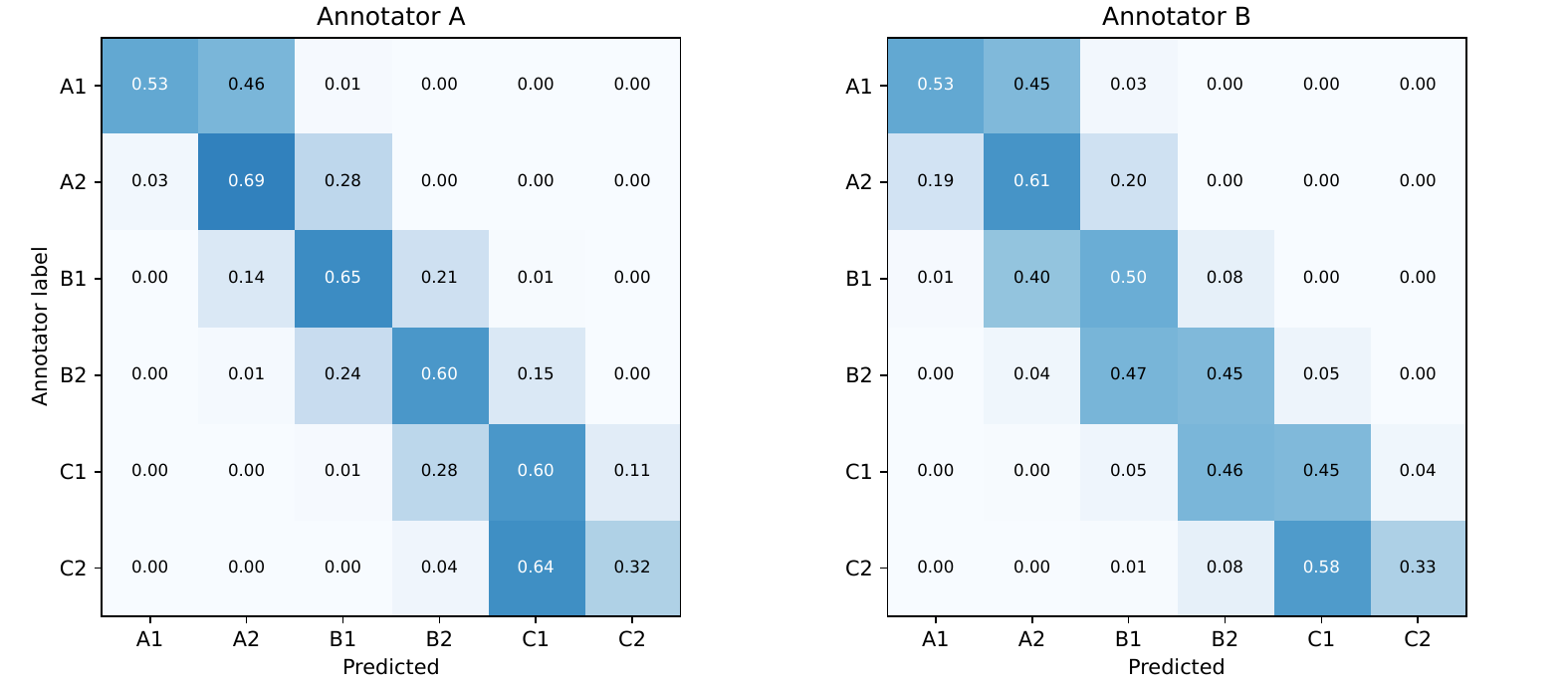} 
  \caption{Confusion matrices of our model on the CEFR-SP test set against each annotator's labels. Cells show row-normalised recall, averaged over $12$ training seeds discarding the best- and worst-performing seed by macro-F1.} 
  \label{fig:confusion-matrix-cefr-sp}
\end{figure*}

\subsection{Evaluation Metrics}
Since the label distributions of all CEFR datasets are highly imbalanced, we report macro-F1, which weights all levels equally and thus prevents the scarce levels from being dominated by the well-represented ones.
For CEFR-SP, where each sentence carries two annotator labels, we count a prediction as correct if it matches either label \cite{arase:2022}.
We report the average scores of over $12$ random seeds for CEFR-SP and $3$ seeds per configuration for UniversalCEFR.

\subsection{Implementation \& Baselines}
We compare our method against two baselines: (i) the same architecture as our method yet trained end-to-end with a plain cross-entropy loss, and (ii) the state-of-the-art on CEFR-SP \cite{arase:2022}, denoted as ``Prototype''.
We employ \texttt{bert-base-uncased} on CEFR-SP and \texttt{xlm-roberta-base} \cite{conneau:2020} on UniversalCEFR as the common PMLM.
Training and hyperparameter settings are in Appendix~\ref{app:implementation}.

\subsection{Results and Discussions}
Table~\ref{tab:main-results} reports per-level F1 and macro-F1 scores on test sets of CEFR-SP and UniversalCEFR.
Overall, \textbf{the proposed method performs best across sentence and document level assessment and monolingual and multilingual settings.}
On CEFR-SP, plain cross-entropy training struggles on the scarce classes A1 and C2, where it exhibits both low scores and high variance across seeds. 
In contrast, the prototype classifier \cite{arase:2022} and our method both mitigate
the class imbalance, improving macro-F1 by more than $12$ points while
remaining on par with cross-entropy on the well-represented intermediate
levels.
On UniversalCEFR, the prototype classifier in turn falls behind cross-entropy at almost all levels, on both granularities.
The proposed method consistently achieves the best macro-F1 scores on all settings.


\begin{table}[t]
\centering
\resizebox{\linewidth}{!}{
\begin{tabular}{lccc}
\toprule
Dataset & Cross-Entropy & Prototype & Proposed \\
\midrule
CEFR-SP & $-0.033$ \tiny{$\pm0.014$} & $-0.014$ \tiny{$\pm0.018$} & $\mathbf{0.011}$ \tiny{$\pm0.005$} \\
UCEFR-sent & $0.045$ \tiny{$\pm0.016$} & $\mathbf{0.116}$ \tiny{$\pm0.011$} & $0.115$ \tiny{$\pm0.014$} \\
UCEFR-doc & $0.070$ \tiny{$\pm0.004$} & $0.168$ \tiny{$\pm0.014$} & $\mathbf{0.190}$ \tiny{$\pm0.014$} \\
\bottomrule
\end{tabular}
}
\caption{Silhouette scores of the test-set embedding spaces with respect to the gold CEFR levels, averaged over the same seeds as Table~\ref{tab:main-results}.
For CEFR-SP, the score is the mean over the two annotators' label sets.}
\label{tab:silhouette}
\end{table}

Moreover, \textbf{supervised contrastive learning produces a more class-separable embedding space.}
Table~\ref{tab:silhouette} reports the silhouette scores \cite{rousseeuw:1987} of the final models' test-set embeddings (visualized in Appendix~\ref{app:tsne}).
The proposed method scores highest on CEFR-SP and at the document level UniversalCEFR, and is on par with the prototype classifier at the sentence level. 


Our analysis found that \textbf{almost all misclassifications occur between adjacent CEFR levels.}
Figure~\ref{fig:confusion-matrix-cefr-sp} shows the confusion matrices of our proposed method, evaluated on CEFR-SP against each annotator's labels.
Errors are concentrated almost exclusively in neighboring classes: $99.1\%$ of test sentences are predicted within one level of the target for Annotator A, and $97.3\%$ for Annotator B.
These results indicate that the proposed method effectively separates distant CEFR levels than baselines but still struggles to distinguish adjacent levels, whose boundaries are inherently ambiguous. 
This observation is consistent with previous findings \cite{arase:2022,imperial:2025}. 
Although the proposed training objective improves class discrimination, it does not explicitly model the ordinal relationships among CEFR levels. 
Incorporating ordinal-aware objectives that preserve the inherent similarity between neighboring levels while improving their discriminability is a promising direction for future work.

\section{Conclusion}
In this paper, we presented a unified training objective for CEFR level assessment that combines focal loss and supervised contrastive learning to address two fundamental challenges of the task: the naturally imbalanced distribution of CEFR levels and the ambiguity between adjacent levels. Extensive experiments on sentence- and document-level CEFR assessment in both monolingual and multilingual settings demonstrated that the proposed method consistently outperforms strong baselines.

The results suggest that jointly addressing class imbalance and adjacent-level ambiguity leads to more discriminative representations for CEFR level prediction and generalizes well across languages and text granularities. As future work, we plan to investigate training objectives that explicitly exploit the ordinal structure of CEFR levels and to extend our approach to other educational NLP tasks involving ordinal difficulty level prediction.

\section*{Limitations}
Although the proposed training objective consistently improves CEFR level assessment across multiple datasets and evaluation settings, several limitations remain. 
First, the method treats CEFR level prediction as a classification task and does not explicitly model the ordinal relationships among levels. 
While supervised contrastive learning learns more discriminative representations, it does not distinguish between adjacent and distant CEFR levels in the optimization objective.

In addition, although we evaluate the proposed method in both monolingual and multilingual settings, we do not investigate its generalization to domains beyond the available benchmarks. 
Evaluating the method on a broader range of educational resources, particularly expert domains such as law and medical, remains an important direction for future work.






\bibliography{custom}

\newpage

\appendix

\section{Dual-Label Handling}
\label{sec:dual_label}
Unlike standard classification datasets, each sentence in CEFR-SP is
annotated by two annotators, resulting in two labels $y_i^{(a)}$ and
$y_i^{(b)}$ per sample $i$. To adapt supervised contrastive learning
to this dual-label setting, we redefine the set of positives $P(i)$
such that a sample $p$ is positive if any of its annotator labels
matches any of $i$'s annotator labels:
\begin{align}
P(i) = \left\{ p \in A(i) \;\middle|\;
\begin{aligned}
& y_i^{(a)} = y_p^{(a)} \;\vee \\
& y_i^{(a)} = y_p^{(b)} \;\vee \\
& y_i^{(b)} = y_p^{(a)} \;\vee \\
& y_i^{(b)} = y_p^{(b)}
\end{aligned}
\right\}
\end{align}
In other words, $i$ and $p$ form a positive pair if they agree on the
CEFR level under at least one combination of their respective
annotators' labels.

\section{Dataset Statistics}
\label{app:dataset-stats}
Table~\ref{tab:dataset-distribution} reports the number of texts per CEFR level for all datasets and splits used in our experiments.

\begin{table*}[t!]
\centering
\small
\begin{tabular}{llrrrrrrr}
\toprule
Dataset & Split & A1 & A2 & B1 & B2 & C1 & C2 & Total \\
\midrule
CEFR-SP (annotator A)
 & train & $513$ & $2{,}772$ & $6{,}329$ & $3{,}603$ & $732$ & $41$ & $13{,}990$ \\
 & dev   & $121$ & $442$ & $680$ & $439$ & $125$ & $36$ & $1{,}843$ \\
 & test  & $109$ & $440$ & $722$ & $419$ & $119$ & $34$ & $1{,}843$ \\
\midrule
CEFR-SP (annotator B)
 & train & $77$ & $1{,}718$ & $5{,}826$ & $4{,}727$ & $1{,}570$ & $72$ & $13{,}990$ \\
 & dev   & $45$ & $207$ & $732$ & $588$ & $222$ & $49$ & $1{,}843$ \\
 & test  & $43$ & $216$ & $715$ & $595$ & $225$ & $49$ & $1{,}843$ \\
\midrule
UniversalCEFR sentence
 & train & $1{,}430$ & $13{,}962$ & $15{,}742$ & $4{,}891$ & \multicolumn{2}{c}{$3{,}448$} & $39{,}473$ \\
 & dev   & $173$ & $2{,}412$ & $2{,}592$ & $586$ & \multicolumn{2}{c}{$383$} & $6{,}146$ \\
 & test  & $247$ & $329$ & $362$ & $249$ & \multicolumn{2}{c}{$213$} & $1{,}400$ \\
\midrule
UniversalCEFR document
 & train & $4{,}851$ & $7{,}446$ & $6{,}702$ & $3{,}838$ & \multicolumn{2}{c}{$2{,}700$} & $25{,}537$ \\
 & dev   & $1{,}578$ & $1{,}752$ & $1{,}516$ & $802$ & \multicolumn{2}{c}{$543$} & $6{,}191$ \\
 & test  & $83$ & $272$ & $406$ & $282$ & \multicolumn{2}{c}{$198$} & $1{,}241$ \\
\bottomrule
\end{tabular}
\caption{Number of texts per CEFR level. Every CEFR-SP sentence is labeled by two annotators. For UniversalCEFR the C1 and C2 classes are merged into a single C class, the merged count spans the C1/C2 columns.}
\label{tab:dataset-distribution}
\end{table*}

\section{Implementation Details}
\label{app:implementation}
All models are optimized using AdamW \cite{loshchilov:2018}, and training was terminated by early stopping (patience of $10$ epochs). 
The batch size was $128$.
For our two-stage approach, the stopping criterion was the silhouette score \cite{rousseeuw:1987} of the resulting embedding space during the first stage, and macro-F1 on the development set during the second stage. 
Hyperparameters were tuned to maximize the macro-F1 scores on the development set (Table~\ref{tab:hyperparams}).

\begin{table}[t]
\centering
\resizebox{\linewidth}{!}{ 
\begin{tabular}{lccc}
\toprule
Hyperparameter & Proposed & CE & Prototype \\
\midrule
$\tau$   & $0.07$ & -- & -- \\
$\gamma$ & $0.5$  & -- & -- \\
$\alpha$ & $0.4$  & -- & $0.2$ \\
LR         & $1e-5$ / $2e-5$\ & $2e-5$ & $1e-5$ \\
\bottomrule
\end{tabular}
}
\caption{Hyperparameters of the proposed method, cross-entropy (CE) and prototype baselines. For the proposed method, the two learning rates (LR) are for the first and second stages, respectively.}
\label{tab:hyperparams}
\end{table}

\section{Embedding Space Visualization}
\label{app:tsne}
\begin{figure*}[t!]
  \centering
  \begin{subfigure}[b]{0.8\textwidth}
    \centering
    \includegraphics[width=\linewidth]{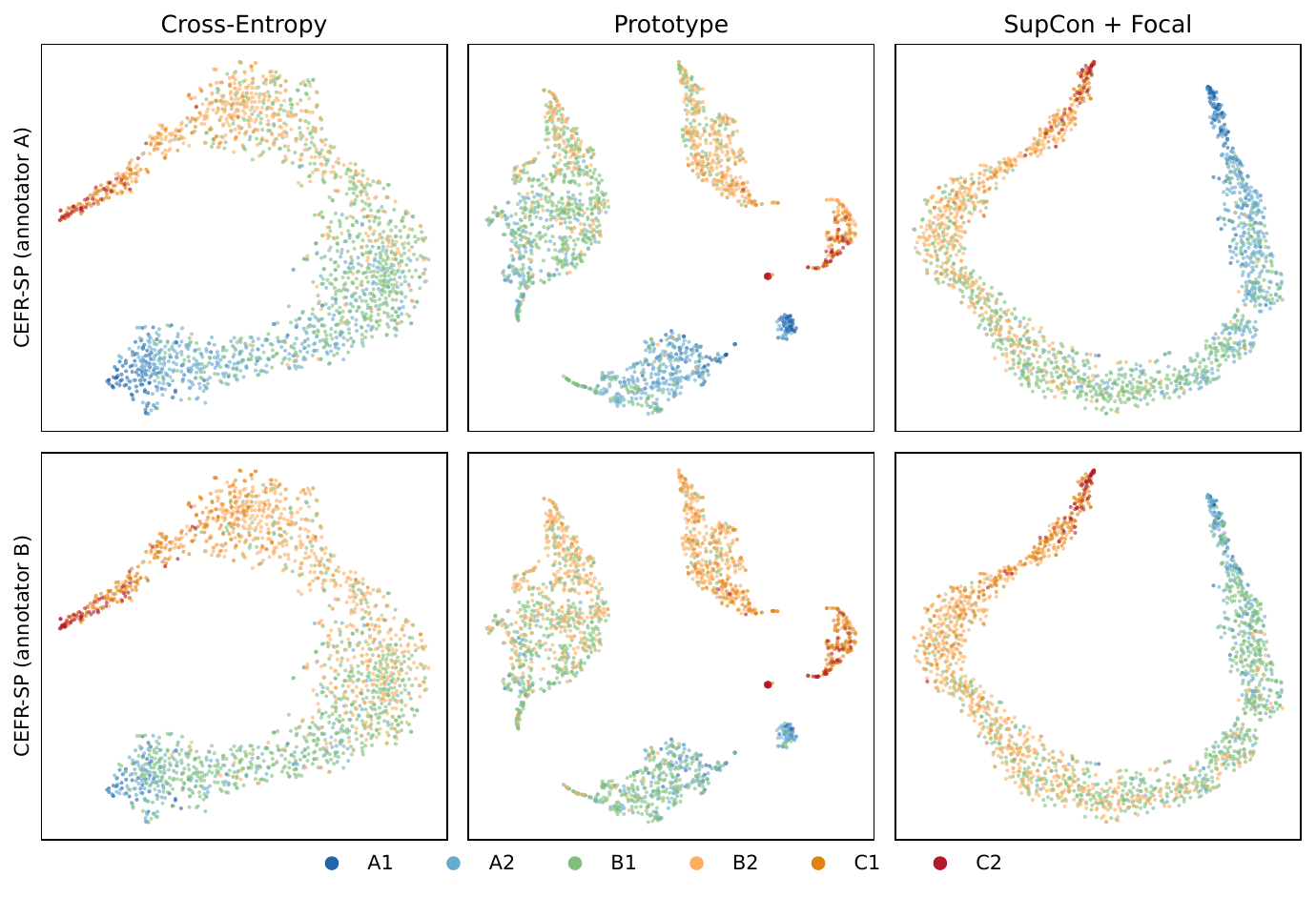}
    \caption{CEFR-SP. Both rows show the same embeddings, colored by annotator A's and annotator B's labels, respectively.}
    \label{fig:tsne-cefrsp}
  \end{subfigure}

  \vspace{0.8em}
  \begin{subfigure}[b]{0.8\textwidth}
    \centering
    \includegraphics[width=\linewidth]{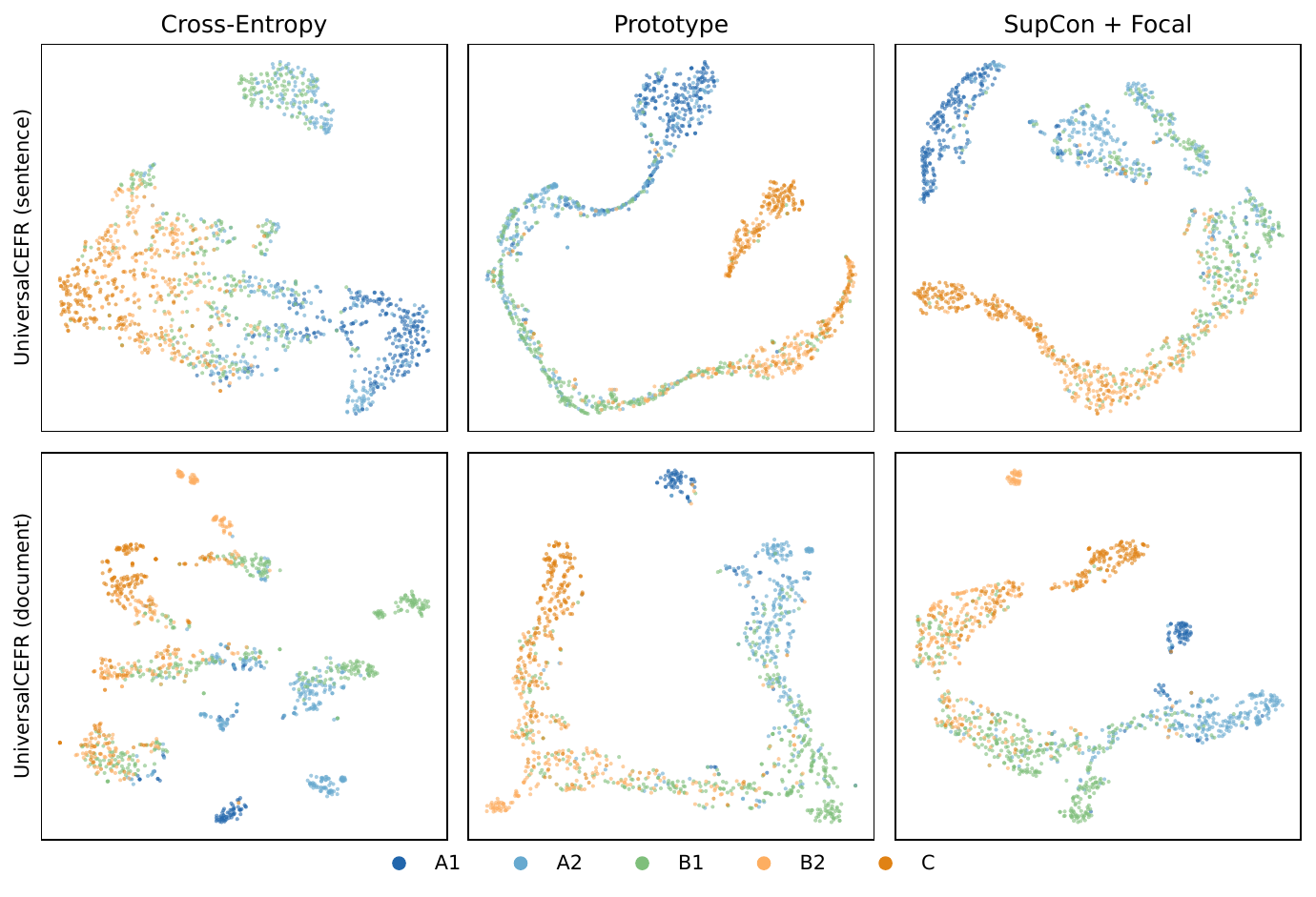}
    \caption{UniversalCEFR, sentence level (top) and document level (bottom), with C1 and C2 merged.}
    \label{fig:tsne-universal-cefr}
  \end{subfigure}
  \caption{t-SNE projections of the test-set embeddings for each method (columns), colored by gold CEFR level. Each panel shows a single training seed: the $6$th-best of $12$ seeds by test macro-F1 for CEFR-SP and the $2$nd-best of $3$ for UniversalCEFR. Panels are independent projections, so distances are not comparable across panels.}
  \label{fig:tsne-embeddings}
\end{figure*}

Figure~\ref{fig:tsne-embeddings} shows t-SNE projections \cite{vandermaaten:2008} of the test-set embedding spaces, for one representative training seed per panel.
On CEFR-SP, the proposed method arranges the levels along a largely ordinal progression.
On UniversalCEFR, it instead separates the levels into multiple clusters, some dominated by a single level and others exhibiting the same ordinal progression.
In both datasets, overlapping levels are almost exclusively adjacent, in line with our observation that misclassifications concentrate on neighboring levels.

\end{document}